\documentclass[twoside,11pt]{article}

\usepackage{dnd}
\usepackage{times}
\usepackage{comment}

\usepackage{comment}
\usepackage{graphicx}
\usepackage{array}
\usepackage{amsmath}
\usepackage{todonotes}
\usepackage{subfig}
\usepackage{graphicx}
\usepackage{soul,xcolor,appendix}
\usepackage{breakurl}
\usepackage[breaklinks]{hyperref}
\definecolor{ForestGreen}{RGB}{34,139,34}
\newsavebox\imagebox 

\dndheading{issue(number)}{year}{firstpage--lastpage}{Name1 Surname1, Name2 Surname2, and Name3 Surname3}{10.5087/dad.DOINUMBER}

\ShortHeadings{Reason and Defenses against Overstability and Oversensitivity of AES Systems}{Singla, Parekh, Singh, Li, Shah and  Chen}
\firstpageno{1}

\begin{document}

\title{Automatic Essay Scoring Systems Are Both Overstable And Oversensitive: Explaining Why And Proposing Defenses}

\author{\name Yaman Kumar Singla\thanks{Equal Contribution} \email ykumar@adobe.com \\
       \addr Adobe Media Data Science Research, IIIT-Delhi, SUNY at Buffalo
               \AND
       \name Swapnil Parekh$^*$ \email swapnil.parekh@nyu.edu \\
       \addr New York University
                \AND 
       \name Somesh Singh$^*$  \email f20180175@goa.bits-pilani.ac.in\\
       \addr IIIT-Delhi
                \AND
       \name Junyi Jessy Li  \email jessy@austin.utexas.edu\\
       \addr University of Texas at Austin
              \AND
       \name Rajiv Ratn Shah  \email rajivratn@iiitd.ac.in\\
       \addr IIIT-Delhi
              \AND
       \name Changyou Chen  \email changyou@buffalo.edu\\
       \addr SUNY at Buffalo
       }

\editor{Name Surname}
\submitted{09/2021}{MM/YYYY}{MM/YYYY}

\maketitle

\begin{abstract}%
 Deep-learning based Automatic Essay Scoring (AES) systems are being actively used by states and language testing agencies alike to evaluate millions of candidates for life-changing decisions ranging from college applications to visa approvals. However, little research has been put to understand and interpret the black-box nature of deep-learning based scoring algorithms. Previous studies indicate that scoring models can be easily fooled. In this paper, we explore the reason behind their surprising adversarial brittleness.  
We utilize recent advances in interpretability to find the extent to which features such as coherence, content, vocabulary, and relevance are important for automated scoring mechanisms. We use this to investigate the oversensitivity ({\it i.e.}, large change in output score with a little change in input essay content) and overstability ({\it i.e.}, little change in output scores with large changes in input essay content) of AES. Our results indicate that autoscoring models, despite getting trained as ``end-to-end'' models with rich contextual embeddings such as BERT, behave like bag-of-words models. A few words determine the essay score without the requirement of any context making the model largely \textit{overstable}. This is in stark contrast to recent probing studies on pre-trained representation learning models, which show that rich linguistic features such as parts-of-speech and morphology are encoded by them. Further, we also find that the models have learnt dataset biases, making them \textit{oversensitive}. The presence of a few words with high co-occurence with a certain score class makes the model associate the essay sample with that score. This causes score changes in $\sim$95\% of samples with an addition of only a few words. To deal with these issues, we propose detection-based protection models that can detect oversensitivity and overstability causing samples with high accuracies. We find that our proposed models are able to detect unusual attribution patterns and flag adversarial samples successfully. 
\end{abstract}


\section{Introduction}
\label{sec:Introdcution}
Automatic Essay Scoring (AES) systems are used in diverse settings such as to alleviate the workload of teachers, save time and costs associated with grading, and to decide admissions to universities and institutions. On average, a British teacher spends 5 hours in a calendar week scoring exams and assignments \citep{micklewright2014teachers}. This figure is even higher for developing and low-resource countries where the teacher to student ratio is dismal. While on the one hand, autograding systems effectively reduce this burden, allowing more working hours for teaching activities, on the other, there have been many complaints against these systems for not scoring the way they are supposed to \citep{flawedAlgos,roboGrade,emptyDream,midDay,perelmanBableWebsite}. For instance, on the recently released automatic scoring system for the state of Utah, students scored lower by writing question-relevant keywords but higher by including unrelated content \citep{flawedAlgos,roboGrade}. Similarly, it has been a common complaint that AES systems focus unjustifiably on obscure and difficult vocabulary \citep{perelmanBable}. While earlier, each score generated by the AI systems was verified by an expert human rater, it is concerning to see that now they are scoring a majority of essays independently without any intervention by human experts \citep{ohioAES2}. The concerns are further alleviated by the fact that the scores awarded by these systems are used in life-changing decisions ranging from college and job applications to visa approvals \citep{greInterpret,educational2014snapshot,UtahNumbers,OhioNumbers}. 

Traditionally, autograding systems are built using manually crafted features used with machine learning based models \citep{kumar2019get}. Lately, these systems have been shifting to deep learning based models \citep{ke2019automated}. For instance, many companies have started scoring candidates using deep learning based automatic scoring \citep{sltiAIRated,trueNorthSA,duolingoEngTest,laflair2019duolingo,yu2015using,chen2018end,singla2021speakerconditioned,riordan2017investigating,pearsonWhitepaper}. However, there are very few research studies on the reliability\footnote{A reliable measure is one that measures a construct consistently across time, individuals, and situations \citep{interspeechTutorialETS}} and validity\footnote{A valid measure is one that measures what it is intended to measure \citep{interspeechTutorialETS}} of ML-based AES systems. More specifically, we have tried to address the problems of robustness and validity which plague deep learning based black-box AES models. Simply measuring test set performance may mean that the model is right for the wrong reasons. Hence, much research is required to understand the scoring algorithms used by AES models and to validate them on linguistic and testing criteria. Similar opinions are expressed by \citet{madnani2018automated} in their position paper on automatic scoring systems.

With this in view, in this paper, we make the following contributions towards understanding current AES systems:

1) Several research studies have shown that essay scoring models are \emph{overstable} \citep{yoon2018atypical,powers2002stumping,kumar2020calling, pham2018path}. Even large changes in essay content do not lead to significant change in scores. For instance, \citet{kumar2020calling} showed that even after changing 20\% words of an essay, the scores do not change much. 
We extend this line of work by addressing \textit{why} the models are overstable. Extending these studies further (\S\ref{sec:AES Overstability}), we investigate AES overstability from the perspective of context, coherence, facts, vocabulary, length, grammar and word choice. We do this by using integrated gradients (\S\ref{sec:attribution-mechanism}), where we find and visualize the most important words for scoring an essay \citep{sundararajan2017axiomatic}. We find that the models despite using rich contextual embeddings and deep learning architectures, are essentially behaving as bag-of-words models.  
Further, with our methods, we also able to improve the adversarial attack strength (\S\ref{sec:Iteratively Deleting Unimportant Words}). For example, for memory networks scoring model \citep{zhao2017memory}, we delete 40\% words from essays without significantly changing score (\textless 1\%), whereas \citet{kumar2020calling} observed that deleting a similar number of words resulted in a decrease of 20\% scores for the same model.


2) While there has been much work on AES overstability \citep{kumar2020calling,perelman2014state,powers2001stumping}, there has been little work on AES oversensitivity. Building on this research gap, by using adversarial triggers, we find that the AES models are also oversensitive, {\it i.e.}, small changes in an essay can lead to large change in scores (\S\ref{sec:AES Oversensitivity}). We find that, by just adding 3 words in an essay containing 350 words (\textless1\% change), we are able to change the predicted score by 50\% (absolute). We explain the oversensitivity of AES systems using integrated gradients \citep{sundararajan2017axiomatic}, a principled tool to discover the importance of parts of an input. The results show that the trigger words added to an essay get unusually high attribution. Additionally, we find the trigger words have usually high co-occurence with certain score labels, thus indicating that the models are relying on spurious correlations causing them to be oversensitive (\S\ref{sec:trigger analysis}). We validate both the oversensitive and overstable sample in a human study (\S\ref{sec:human baseline}). We ask the annotators whether the scores given by AES models are right by providing them with both original and modified essay responses and scores.


3) While much previous research in the linguistic field studies how essay scoring systems can be fooled, for the first time, we propose models that can detect samples causing overstability and oversensitivity \citep{pham2020out,kumar2020calling,perelman2020}. Our models are able to detect both overstability and oversensitivity causing samples with high accuracies (\textgreater 90\% in most cases) (\S\ref{sec:Proposed Solutions For Detection Of Oversensitive and Overstable Sample}). Also, for the first time in the literature, through these solutions, we propose a simple yet effective solution for universal adversarial peturbation (\S\ref{sec:IG Based Oversensitive Sample Detection}). These models, apart from defending AES systems against oversensitivity and overstability causing samples, can also inform effective human intervention strategy. For instance, AES deployments either completely rely on double scoring essay samples (human and machine) or solely on machine ratings alone \citep{etsDulingoComparison} can choose to have an effective middle ground. Using our models, AES deployments can select samples for human testing and intervention more effectively. Public school systems, \textit{e.g.}, in Ohio which use automatic scoring without any human interventions can select samples using these models for limited human intervention \citep{ohioAES2,OhioNumbers}. For this, we also conduct a small-scale pilot study on the AES deployment of a major language testing company proving the efficacy of the system (\S\ref{sec:pilot study}). Previous solutions for human interventions optimization rely on brittle features such as number of words and content modeling approaches like off-topic detection \citep{yoon2018atypical,yoon2017combining}. These models cannot detect adversarial samples like the ones we present in our work. 


We perform our experiments for three model architectures and eight unique prompts\footnote{Here a prompt denotes an instance of a unique question asked to test-takers for eliciting their opinions and answers in an exam. The prompts can come from varied domains including literature, science, logic and society. The responses to a prompt indicate the creative, literary, argumentative, narrative, and scientific aptitude of candidates and are judged on a pre-determined score scale.}, demonstrating the results on twenty four unique model-dataset pairs. It is worth noting that our goal in this paper is \textbf{not} to show that AES systems can be fooled \textit{easily}. Rather, \textbf{our goal is to interpret how deep-learning based scoring models score essays, why they are overstable and oversensitive, and how to solve the problems of oversensitivity and overstability.}
We release all our code,
dataset and tools for public use with the hope that it will spur testing and validation of AES models considering their huge impact on the lives of millions every year 
\citep{educational2014snapshot}.


\section{Background}
\label{background}

\subsection{Task, Models and Dataset}
\label{task-and-dataset}
We use the widely cited \citet{ASAP-AES} dataset which comes from Kaggle Automated Student Assessment Prize (ASAP) 
for the evaluation of automatic essay scoring systems. The ASAP-AES dataset has been used for automatically scoring essay responses by many research studies \citep{taghipour2016neural,easeGithub,tay2018skipflow}. It is one of the largest publicly available datasets (Table~\ref{table:AES-dataset-stats}). The questions covered by the dataset span many different areas such as Sciences and English. The responses were written by high school students and were subsequently double-scored. 

\begin{table*}[htbp]
\small
\centering
\resizebox{\textwidth}{!}{\begin{tabular}{lllllllll}
\hline
\textbf{Prompt Number} & \textbf{1} & \textbf{2} & \textbf{3} &\textbf{ 4} & \textbf{5}  &\textbf{ 6 } &\textbf{ 7 } & \textbf{8}  \\\hline
\#Responses & 1783 & 1800 & 1726 & 1772 & 1805 & 1800 & 1569 & 723  \\
Score Range & 2-12 & 1-6  & 0-3  & 0-3  & 0-4  & 0-4  & 0-30 & 0-60 \\
\#Avg words per response & 350 & 350  & 150  & 150  & 150  & 150 & 250 & 650  \\
\#Avg sentences per response & 23 &20 &6 &4 &7 &8 &12 &35   \\ 
Type & Argumentative & Argumentative &RC &RC &RC &RC & Narrative  &Narrative\\ \hline
\end{tabular}}
\caption{\label{table:AES-dataset-stats} Overview of the ASAP AES Dataset used for evaluation of AS systems. \small (RC = Reading Comprehension). 
}
\end{table*}

We test the following two state-of-the-art models in this work: \textit{SkipFlow} \citep{tay2018skipflow} and Memory Augmented Neural Network (\emph{MANN}) \citep{zhao2017memory}. Further, for comparison, we design a BERT based automatic scoring model. 
The performance is measured using Quadratic Weighted Kappa (QWK) metric, which indicates the agreement between a model's and the expert human rater's scores. All models show an improvement of 4-5\% over the previous models on the QWK metric. The analysis of these models, especially BERT, is interesting in light of recent studies indicating that pre-trained language models learn rich linguistic features including morphology, parts-of-speech, word-length, noun-verb agreement, coherence, and language delivery \citep{conneau2018you,hewitt2019structural,shah2021all}. This has resulted in pushing the envelope for many NLP applications. The individual models we use are briefly explained as follows:

\paragraph{SkipFlow} 
\citet{tay2018skipflow} model essay scoring as a regression task. They utilize Glove embeddings for representing the tokens. SkipFlow captures coherence, flow and semantic relatedness over time, which
the authors call neural coherence features. 
Due to the intelligent modelling, it gets an impressive average quadratic weighted kappa score of 0.764. SkipFlow is one of the top performing models \citep{tay2018skipflow,ke2019automated} for AES.

\paragraph{MANN} \citet{zhao2017memory} use memory networks for autoscoring by selecting some responses for each grade. These responses
are stored in memory and then used for scoring ungraded
responses. The memory component helps to characterize the
various score levels similar to what a rubric does. They show an excellent agreement score of 0.78 average QWK outperforming the previous state-of-the-art models.




\paragraph{BERT-based}
We also design a BERT-based architecture for scoring essays. It utilizes BERT embeddings \citep{devlin2018bert} to represent essays by passing tokens through the BERT Encoder. The resultant embeddings from the last layer of the encoder are pooled and passed through a fully connected layer of size 1 to produce the score. The network was trained to predict the essay scores by minimizing the mean squared error loss. It achieves an average QWK score of 0.74. We also try out finetuning BERT to see its effect on downstream task \citep{mayfield2020should}. It does not result in any statistically significant difference in results. We utilize this architecture as a baseline representative of transformer-based embedding models.

\subsection{Attribution Mechanism}
\label{sec:attribution-mechanism}

The task of attributing a score $F(x)$ given by an AES model $F$, on an input essay $x$ can be formally defined as producing attributions $a_1,..,a_n$ corresponding to the words $w_1,..,w_n$ contained in the essay $x$. The attributions produced are such that\footnote{Proposition 1 in \citep{sundararajan2017axiomatic}} $Sum(a_1,..,a_n) = F(x)$, \emph{i.e.,} net attributions of all words ($Sum(a_1,..,a_n)$)
equal the assigned score ($F(x)$). In a way, if $F$ is a regression based model, $a_1,..,a_n$ can be thought of as the scores of each word of that essay, which sum to produce the final score, $F(x)$.

We use a path-based attribution method, Integrated Gradients (IGs) \citep{sundararajan2017axiomatic}, much like other interpretability mechanisms such as \citep{ribeiro2016lime, lundberg2017shap} for getting the attributions for each of the trained models, $F$. IGs employ the following method to find blame assignments:
given an input $x$ and a baseline
$b$\footnote{Defined as an input containing absence of cause for the output of a model; also called neutral input \citep{shrikumar2016not,sundararajan2017axiomatic}.}, the integrated gradient along the $i^{th}$ dimension is defined as:
\begin{equation}
\label{eqn:intgrad}
IG_i(x,b) = (x_i-b_i)\int_{\alpha=0}^{1} \frac{\partial F(b + \alpha(x-b))}{\partial x_i  }~d\alpha
\end{equation}
where $\frac{\partial F(x)}{\partial x_i}$ represents the gradient of $F$ along the $i^{th}$ dimension of  $x$.

\begin{figure*}[!htbp]

 \centering
 \begin{tabular}{c}

 \includegraphics[scale = 0.73]{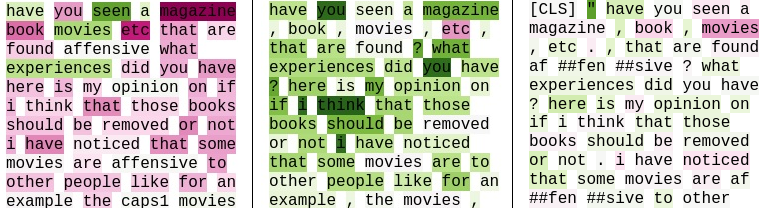}\\

 \end{tabular}
 \caption{\small Attributions for SkipFlow, MANN and BERT models respectively of an essay sample for Prompt 2. Prompt 2 asks candidates to write an essay to a newspaper reflecting their views on censorship in libraries and express their views if they believe that materials, such as books, \textit{etc.}, should be removed from the shelves if they are found offensive. This essay scored 3 out of 6. 
 }
 \label{fig:Skipflow and MANN Normal}
\end{figure*}

We choose the baseline as empty input (all 0s) for essay scoring models since an empty essay should get a score of 0 as per the scoring rubrics. It is the neutral input that models the absence of a cause of any score, thus getting a zero score. Since we want to see the effect of only words on the score, any additional inputs (such as memory in MANN) of the baseline $b$ is set to be that of $x$\footnote{We ensure that IGs are within the acceptable error margin of \textless 5\%, where the error is calculated by the property that the attributions' sum should be equal to the difference between the probabilities of the input and the baseline. IG parameters: Number of Repetitions = 20-50, Internal Batch Size = 20-50}. See Fig.~\ref{fig:Skipflow and MANN Normal} for an example.

We choose IGs over other explainability techniques since they have many desirable properties that make them useful for this task. For instance, the attributions sum to the score of an essay ($Sum(a_1,..,a_n)$ = $F(x)$), they are implementation invariant, do not require any model to be retrained and are readily implementable. Previous literature such as \citep{mudrakarta2018did} also uses Integrated Gradients for explaining the undersensitivity of factoid based question-answer (QA) models. Other interpretability mechanisms like attention require changes in the tested model and are not post-hoc, thus are not a good choice for our task.


\section{Empirical Studies and Results}
\label{sec:experiments and results}

We perform our overstability (\S\ref{sec:AES Overstability}) and oversensitivity (\S\ref{sec:AES Oversensitivity}) experiments with 100 samples per prompt for the three models discussed in Section~\ref{task-and-dataset}. There are 8 prompt-level datasets in the overall ASAP-AES dataset, therefore we perform our analysis on 24 unique model-dataset pairs, each containing over 100 samples.


\subsection{AES Overstability}
\label{sec:AES Overstability}
 We first present results on model overstability. Following the previous studies, we test the models' overstability on different features important for AES scoring such as the knowledge of context (\S\ref{sec:Attribution of original samples}, \ref{sec:Iteratively Deleting Unimportant Words}), coherence (\S\ref{sec:Shuffling of sentences}), facts (\S\ref{sec:Knowledge of Factuality}), vocabulary (\S\ref{sec:Modification in Lexicon}), length (\S\ref{sec:Iteratively Deleting Unimportant Words}), meaning (\S\ref{sec:Shuffling of sentences}), and grammar (\S\ref{sec:Modification in Lexicon}). This set of features provides an exhaustive coverage of all features important for scoring essays \citep{yan2020handbook}.

\subsubsection{Attribution of original samples}
\label{sec:Attribution of original samples}
We take the original human-written essays from the ASAP-AES dataset and do a word-level attribution of scores. Fig.~\ref{fig:Skipflow and MANN Normal} shows the attributions of all models for an essay sample from Prompt 2. We observe that SkipFlow does not attribute any word after the first few lines (first 30\% essay content) of the essay while MANN attributions are spread over the complete length of the essay. For BERT-based model, we see that most of the attributions are over non linguistic features (tokens) like \textit{`CLS'} and \textit{`SEP'}. \textit{CLS} and \textit{SEP} tokens are used as delimiters in the BERT model. A similar result was also observed by \citet{kovaleva2019revealing}.
 
 For SkipFlow, we observe that if a word is negatively attributed at a certain position in an essay sample, it is then commonly negatively attributed in its other occurrences as well. For instance, \textit{books}, \textit{magazines} were negatively attributed in all its occurrences while \textit{materials}, \textit{censored} were positively attributed and \textit{library} was not attributed at all. We could not find any patterns in the direction of attribution. In MANN, the same word changes its attribution sign when present in different essays. However, in a single instance of an essay, a word shows the same sign overall despite occurring in very different contexts.
 
Table~\ref{table:attributed words-normal-samples} lists the top-positive, top-negative attributed words and the mostly unattributed words for all models. For MANN, we notice that the attributions are stronger for function words like, \textit{to, of, you, do,} and \textit{are} and lesser for content words like, \textit{shelves, libraries,} and \textit{music}. SkipFlow's top attributions are mostly construct-relevant words while BERT also focuses more on stop-words. 
 \begin{figure*}[htbp]
 \centering
 \begingroup
\renewcommand{\arraystretch}{1}

 \begin{tabular}{l|l|l}
\includegraphics[scale=0.17]{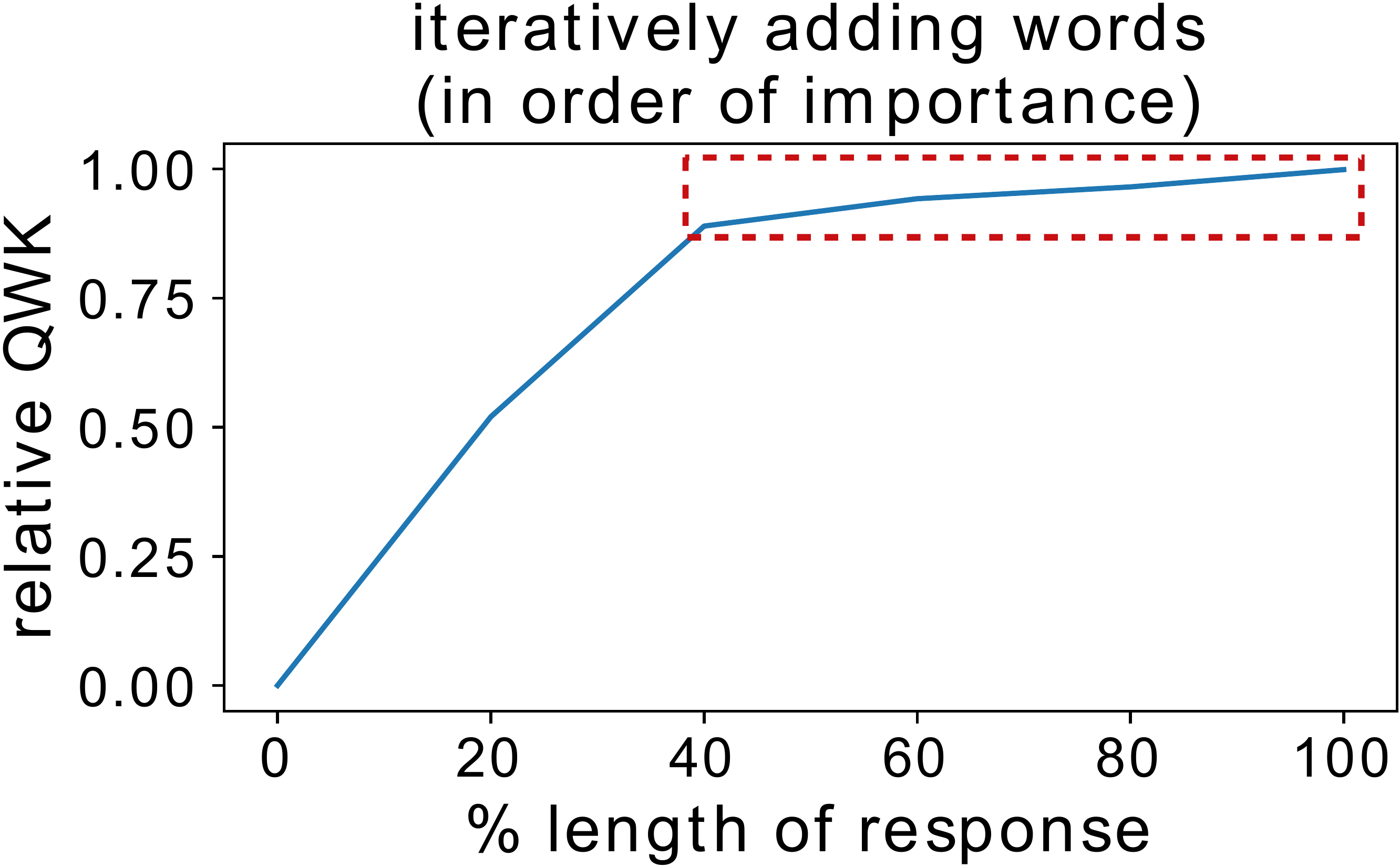} & %
\includegraphics[scale=0.17]{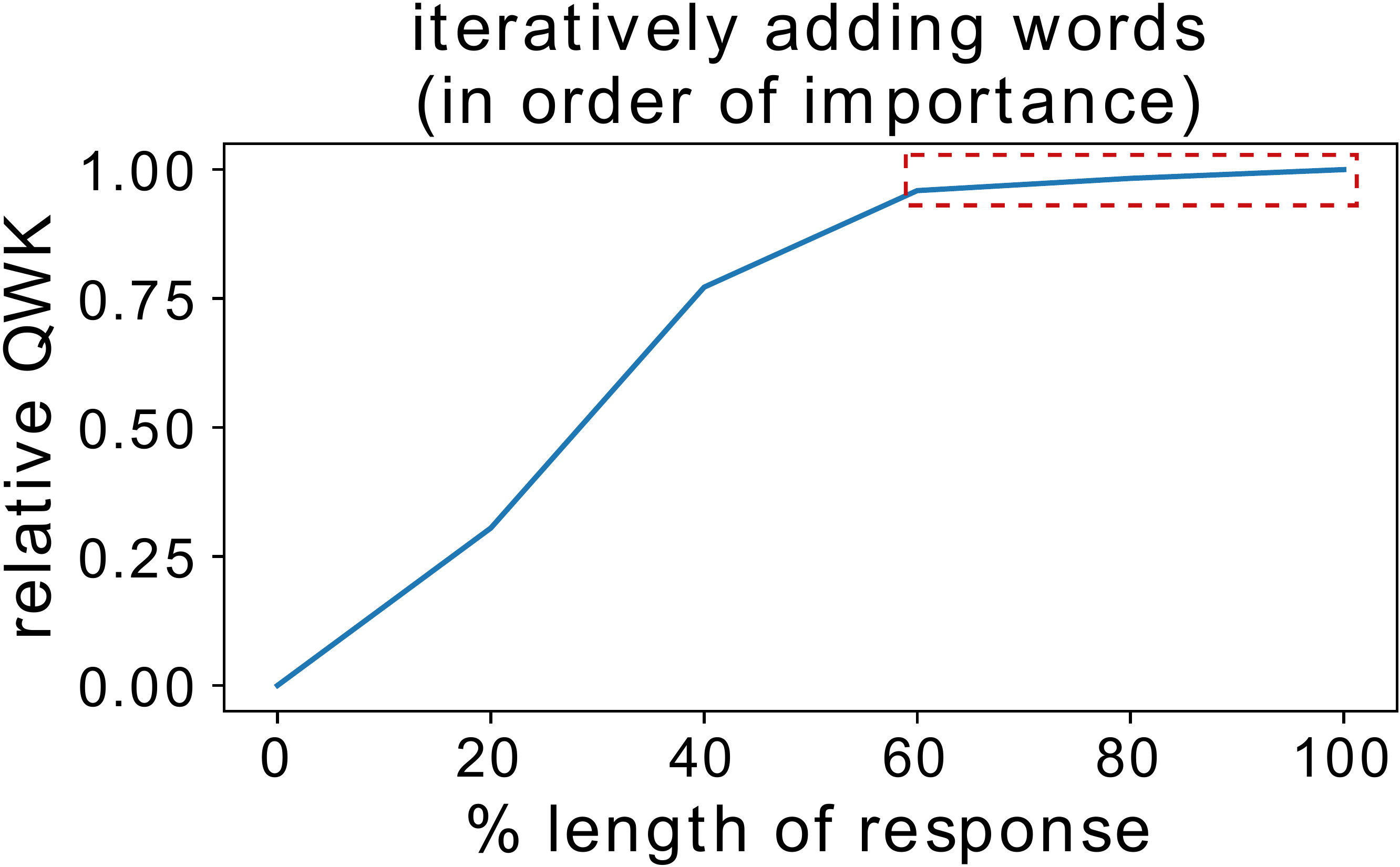} & %
\includegraphics[scale=0.17]{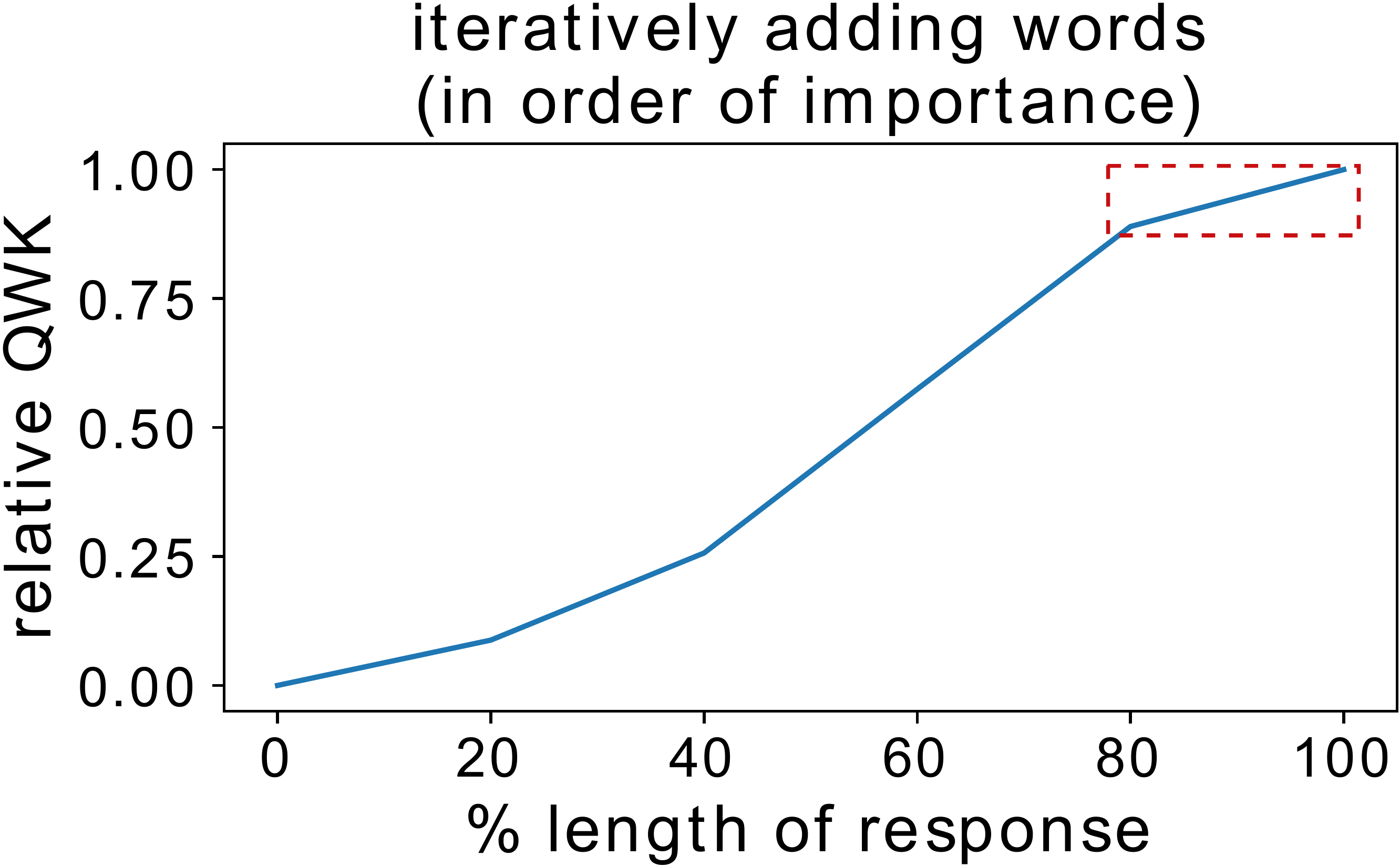} \\ 

 \end{tabular}
\endgroup

 \caption{\small Variation of QWK with iterative addition of response words for SkipFlow, MANN and BERT models. The y-axis notes the relative QWK with respect to the original QWK and the x-axis represents iterative addition of attribute-sorted response words. These results are obtained on Prompt 7, similar results were obtained for all the prompts tested.
 }
 \label{fig:Skipflow and MANN Removing and Adding Words}
\end{figure*}

\subsubsection{Iteratively Deleting Unimportant Words}
\label{sec:Iteratively Deleting Unimportant Words}

 For this test, we take the original samples and iteratively delete the least attributed words (Eq.~\ref{eq:iteratively deleting}).
 \begin{equation}
 \label{eq:iteratively deleting}
 IG-attribution~sorted~list~of~tokens~=~(x_1, x_2, ..., x_k, ....., x_n)  
\end{equation}
such that $IG(x_1,b)>IG(x_2,b)> .. > IG(x_k,b)$, where $x_k$ represents the $k^{th}$ essay token to be removed, b represents the baseline and $IG(x_k,b)$ represents the attribution on $x_k$ with respect to baseline $b$.

 Through this, we note the model's dependence on a few words without their context.
 Fig.~\ref{fig:Skipflow and MANN Removing and Adding Words} presents the results. We observe that the performance (measured by QWK) for the BERT model stays within 95\% of the original performance even if one of every four words was removed from the essays in the reverse order of their attribution values. The percentage of words deleted were even more for the other models. While Fig.~\ref{fig:Skipflow and MANN Normal} showed that MANN paid attention to the full length of the response, removing words does not seem to affect the scores much. Notably, the words removed are not contiguous but interspersed across sentences, therefore deleting the unattributed words does not produce a grammatically correct response (see Fig.~\ref{fig:Skipflow and MANN Word Soup}), yet can get a similar score thus defeating the whole purpose of testing and feedback.

 These findings show that there is a point after which the score flattens out, \emph{i.e.}, it does not change in that region either by adding or removing words. This is odd since adding or removing a word from a sentence typically alters its meaning and grammaticality, yet the models do not seem to be affected; they decide their scores only based on 30-50\% words. As an example, a sample after deleting bottom 40\% attributed words is given here:
``\st{In the end} patience rewards \st{better} than impatience. \st{A time} that \st{I was} patient \st{was} last year \st{at} cheer competition.''

\begin{table}[htbp]
	\centering
	\footnotesize
	\begin{tabular}{|l|l|}
	 \hline	\hline \textbf{Model} & \textbf{Positively Attributed Words} \\ \hline
		MANN & to, of, are, ,, children, do, ', we  \\ \hline
		SKIPFLOW &  of, offensive, movies, censorship, is, our \\ \hline 
		BERT &  ., the, to, and, ", was, caps, [CLS] \\ \hline
		
		\hline \textbf{Model} & \textbf{Negatively Attributed Words} \\ \hline
		MANN & i, shelf, by, shelves, libraries, music, a \\ \hline
		SKIPFLOW & the, i, to, in, that, do, a, or, be \\ \hline 
		BERT &  i, [SEP], said, a, in, time, one \\ \hline 

     \hline \textbf{Model} & \textbf{Mostly Unattributed Words} \\ \hline
        MANN &  t, you, the, think, offensive, from, my \\ \hline
		SKIPFLOW &  it, be, but, their, from, dont, one, what \\ \hline 
		BERT &  @, \#\#1, and, ,, my, patient \\ \hline 
        
	\end{tabular}
	\caption{
		\small
		\label{table:attributed words-normal-samples} 
		Top positive, negative and un-attributed words for SkipFlow, MANN and BERT-based model for Prompt 2.
	}
\end{table}

\subsubsection{Sentence and Word Shuffle} \label{sec:Shuffling of sentences}

\begin{figure*}[htbp]
     \centering
     \begin{tabular}{c}
    \includegraphics[scale=0.70]{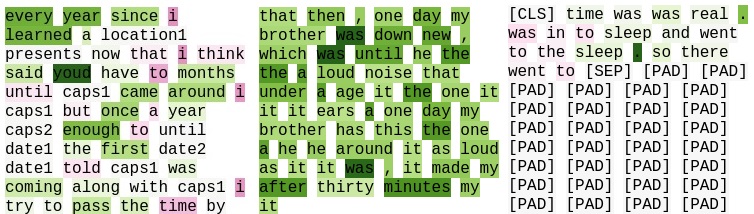} \\
     \end{tabular}

     \caption{\small Word-shuffled essay containing 40\% of (top-attributed) words for SkipFlow (left), MANN (middle) and BERT (right) models respectively. The perturbed essay scores 26 (SkipFlow), 15 (MANN) and 5 (BERT) out of 30. The original essay was scored 25, 16, 4 respectively by the models.
     }
     \label{fig:Skipflow and MANN Word Soup}
\end{figure*}

Coherence and organization are important features for scoring: they measure unity of different ideas in an essay and determine its cohesiveness in the narrative \citep{barzilay2008modeling}. To check the dependence of AES models on coherence, we shuffle the order of sentences and words randomly and note the change in score between the original and modified essay (Fig.~\ref{fig:Skipflow and MANN Word Soup}). 

 We observe little change (\textless 0.002\%) in the attributions with sentence shuffle. The attributions are mostly dependent on word identities rather than their position and context for all models. 
 We also find that shuffling sentences results in 10\%, 2\% and 3\% difference in scores for SkipFlow, MANN, and BERT models, respectively. Even for these samples for which we observed a change in the scores, almost half of them increased their scores and the other half was reduced. The results are similar for word-level shuffling. This is surprising since change in the order of ideas in an essay alters the meaning of a prose but the models are unable to detect changes in either idea order or word-order. It indicates that despite getting trained as sentence and paragraph level models with the knowledge of language models, they have essentially become \textit{bag-of-words models}. 


\subsubsection{Modification in Lexicon}
\label{sec:Modification in Lexicon}

Several previous research studies have highlighted the importance vocabulary plays in scoring and how AES models may be biased towards obscure and difficult vocabulary \citep{perelmanBable,perelman2014state,hesse20052005,powers2002stumping,kumar2020calling}. To verify their claims, we replace the top and bottom 10\% attributed words with `similar' words\footnote{Sampled from Glove with the distance calculated using Euclidean distance metric \citep{pennington-etal-2014-glove}}. 

Table~\ref{table:modify lexicon} shows the results for this test. It can be noted that after replacing all the top and bottom 10\% attributed words with their corresponding `similar' words results in an average 4.2\% difference in scores across all the models. 
These results imply that networks are surprisingly not perturbed by modifying even the most attributed words and produce equivalent results with other similarly-placed words. In addition, while replacing a word with a `similar' word often changes the meaning and form of a sentence\footnote{For example, consider the replacement of the word `agility' with its synonym `cleverness' in the sentence `This exercise requires agility.' does not produce a sentence with the same meaning.}, the models do not recognize that change by showing no change in their scores. 


\begin{table}[htbp]
	\centering
	\footnotesize
	\begin{tabular}{|l|lll|}
	\hline Result & SkipFlow & MANN & BERT \\ \hline
	Avg score difference & 9.8\% & 2.4\% & 3\% \\ \hline
	\% of top-20\% attributed\\words which change\\attribution & 20.3\% & 9.5\% & 34\%\\ \hline 
	 \% of bottom-20\% attributed\\words which change\\attribution & 22.5\% & 26.0\% & 45\% \\ \hline 
	\end{tabular}
	\caption{
		\small
		\label{table:modify lexicon} 
		Statistics obtained after replacing the top and bottom 10\% attributed words of each essay with their synonyms. 
	} 

\end{table}

\begin{figure*}[htbp]
 \centering
 \begin{tabular}{c}
\includegraphics[scale=0.80]{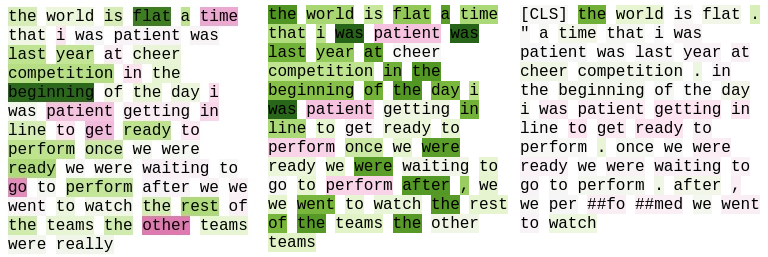}
 \end{tabular}
 
 \caption{\small Attributions for SkipFlow (left), MANN (middle) and BERT (right) models of an essay sample where a false fact has been introduced at the beginning. This essay sample scores (25/30, 18/30, 22/30) by the three models respectively. The original essay (without the added lie) scored (24/30), (18/30) and (21/30) respectively.}

\vspace*{-5 mm}

 \label{fig:Skipflow and MANN Lies}
\end{figure*}


\subsubsection{Factuality, Common Sense and World Knowledge} 
\label{sec:Knowledge of Factuality}

Factuality, common sense and world knowledge are important features in scoring essays \citep{yan2020handbook}. While a human expert can readily catch a lie, it is difficult for a machine to do so. We randomly sample 100 sample essays of each prompt from the \textsc{AddLies} test case of \citep{kumar2020calling}. For constructing these samples, they used various online databases and appended the false information at various positions in the essay. These statements not only introduce false facts in the essay but also perturb its coherence.

 A teacher who is responsible for teaching, scoring, and feedback of a student must have knowledge of world knowledge such as, `Sun rises in the East', and `The world is not flat'. However, Fig.~\ref{fig:Skipflow and MANN Lies} shows that scoring models  
do not have the capability to check such common sense. The models tested in fact attribute positive scores to statements like \emph{the world is flat} if present at the beginning. These results are in contrast with studies like \citep{tenney2019you,zhou2020evaluating} which indicate that BERT and Glove like contextual representations have common sense and world knowledge. \citet{ettinger2020bert} in their `negation test' also observe similar results to us. 

\textbf{BABEL Semantic Garbage:} Linguistic literature has also reported that unexplicably, AES models give high scores to semantic garbage like the one generated using B.S. Essay Language Generator (BABEL generator)\footnote{\url{https://babel-generator.herokuapp.com/}} \citep{perelmanBable,perelmanBableWebsite,perelman2020}. These samples are essentially semantic garbage with perfect spellings and obscure and lexically complex vocabulary. 
In stark contrast to \citep{perelmanBable} and the commonly held notion that writing obscure and difficult to use words fetch more marks, we observed that the models attributed infrequent words such as \emph{forbearance, legerdemain,} and \emph{propinquity} negatively while common words such as \emph{establishment, celebration,} and \emph{demonstration} were positively scored. Therefore, our results show no evidence for the hypothesis reported by studies like \citep{perelman2020} that writing lexically complex words make the AES systems give better scores.

\subsection{AES Oversensitivity}
\label{sec:AES Oversensitivity}

While there has been literature for AES overstability, there is much less literature on AES oversensitivity. Therefore, next using universal adversarial triggers \citep{wallace2019universal}, we show the oversensitivity of AES models. We add a few words (adversarial triggers) to the essays and cause them to have large changes in their scores. Post that, we attribute the oversensitivity to essay words and show that trigger words have high attributions and are the ones responsible for the model oversensitivity.

Through this, we test whether an automatically generated small phrase can perform an untargeted attack against a model to increase or decrease the predicted scores irrespective of the original input. Our results show that these models are vulnerable to such attacks, with as few as three tokens increasing / decreasing the scores of $\approx99\%$ of samples. Further, we show the performance of transfer attacks across prompts and find that $\approx80\%$ of them transfer, thus showing that the adversaries are easily domain adaptable and transfer well across prompts\footnote{For the consideration of space, we only report a subset of these results.}. We choose to use universal adversarial triggers for this task since they are input-agnostic, consist of a small number of tokens, and since they do not require the model's white box access for every essay sample, they have the potential of being used as ``cheat-codes" where a code once extracted can be used by every test-taker. Our results show that the triggers are highly effective.  


\begin{table}[ht]
\begin{tabular}{|c||c|c|c|c|c|}

        Prompt\textrightarrow  & 1 & 4 & 6 & 7 & 8\\ \hline \hline
        Trigger Len\textdownarrow & \multicolumn{5}{c}{Model = SkipFlow} \\ \hline
        3  &  
        \textcolor{ForestGreen}{68},  \textcolor{red}{43} &   \textcolor{ForestGreen}{\bf 100},  \textcolor{red}{14} &   \textcolor{ForestGreen}{86},  \textcolor{red}{40} &   \textcolor{ForestGreen}{56},  \textcolor{red}{81} &   \textcolor{ForestGreen}{43},  \textcolor{red}{75}    \\
        5 &   
        \textcolor{ForestGreen}{79},  \textcolor{red}{38} &    \textcolor{ForestGreen}{\bf 100},  \textcolor{red}{13} &   \textcolor{ForestGreen}{\bf 97},  \textcolor{red}{42} &   \textcolor{ForestGreen}{65},  \textcolor{red}{83} &   \textcolor{ForestGreen}{44},  \textcolor{red}{78}    \\
        10 &   
        \textcolor{ForestGreen}{85},  \textcolor{red}{44} &   \textcolor{ForestGreen}{\bf 100},  \textcolor{red}{18} &   \textcolor{ForestGreen}{\bf 100},  \textcolor{red}{48} &   \textcolor{ForestGreen}{78},  \textcolor{red}{88} &   \textcolor{ForestGreen}{55},  \textcolor{red}{\bf 94}    \\
        20 &   \textcolor{ForestGreen}{\bf 93},  \textcolor{red}{\bf 68} &   \textcolor{ForestGreen}{\bf 100},  \textcolor{red}{27} &   \textcolor{ForestGreen}{\bf 100},  \textcolor{red}{\bf 58} &   \textcolor{ForestGreen}{\bf 90},  \textcolor{red}{\bf 91} &   \textcolor{ForestGreen}{67},  \textcolor{red}{\bf 99}   \\
        \hline \hline

        & \multicolumn{5}{c}{Model = BERT} \\ \hline 
        3 &   \textcolor{ForestGreen}{71},  \textcolor{red}{53} &   \textcolor{ForestGreen}{89},  \textcolor{red}{31} &   \textcolor{ForestGreen}{66},  \textcolor{red}{27} &   \textcolor{ForestGreen}{55},  \textcolor{red}{77} &   \textcolor{ForestGreen}{46},  \textcolor{red}{61}   \\
        5 &   \textcolor{ForestGreen}{77},  \textcolor{red}{52} &    \textcolor{ForestGreen}{\bf 90},  \textcolor{red}{33} &   \textcolor{ForestGreen}{73},  \textcolor{red}{33} &   \textcolor{ForestGreen}{58},  \textcolor{red}{79} &   \textcolor{ForestGreen}{49},  \textcolor{red}{64}    \\
        10 &   \textcolor{ForestGreen}{79},  \textcolor{red}{55} &   \textcolor{ForestGreen}{\bf 91},  \textcolor{red}{41} &   \textcolor{ForestGreen}{87},  \textcolor{red}{48} &   \textcolor{ForestGreen}{68},  \textcolor{red}{84} &   \textcolor{ForestGreen}{55},  \textcolor{red}{75}   \\
        20 &   \textcolor{ForestGreen}{83},  \textcolor{red}{61} &   \textcolor{ForestGreen}{\bf 94},  \textcolor{red}{\bf 49} &   \textcolor{ForestGreen}{\bf 95},  \textcolor{red}{\bf 59} &   \textcolor{ForestGreen}{88},  \textcolor{red}{89} &   \textcolor{ForestGreen}{61},  \textcolor{red}{89}   \\
        \hline \hline

        & \multicolumn{5}{c}{Model = MANN}{}  \\ \hline
        3 &  \textcolor{ForestGreen}{67},  \textcolor{red}{38} &  \textcolor{ForestGreen}{89},  \textcolor{red}{15} &  \textcolor{ForestGreen}{86},  \textcolor{red}{40} &  \textcolor{ForestGreen}{60},  \textcolor{red}{80} &  \textcolor{ForestGreen}{41},  \textcolor{red}{70}\\
        5 &  \textcolor{ForestGreen}{73},  \textcolor{red}{39} &  \textcolor{ForestGreen}{\bf 93},  \textcolor{red}{19} &  \textcolor{ForestGreen}{\bf 96},  \textcolor{red}{42} &  \textcolor{ForestGreen}{61},  \textcolor{red}{71} &  \textcolor{ForestGreen}{43},  \textcolor{red}{77}\\
        10 &  \textcolor{ForestGreen}{85},  \textcolor{red}{44} &  \textcolor{ForestGreen}{\bf 97},  \textcolor{red}{20} &  \textcolor{ForestGreen}{\bf 99},  \textcolor{red}{48} &  \textcolor{ForestGreen}{75},  \textcolor{red}{84} &  \textcolor{ForestGreen}{59},  \textcolor{red}{88}\\
        20 &  \textcolor{ForestGreen}{\bf 93},  \textcolor{red}{63} &  \textcolor{ForestGreen}{\bf 100},  \textcolor{red}{20} &  \textcolor{ForestGreen}{\bf 100},  \textcolor{red}{\bf 59} &  \textcolor{ForestGreen}{84},  \textcolor{red}{\bf 90} &  \textcolor{ForestGreen}{\bf 71},  \textcolor{red}{\bf 94}\\
                        \hline \hline
\end{tabular}
\caption{
		\small
		\label{table:oversensitivity_single}
		Single-prompt targeted attack performance results.  \textcolor{ForestGreen}{Percentage of samples whose scores increase}, \textcolor{red}{Percentage of samples whose scores decrease} on using triggers of length $c$ (Rows, Up to Down) on prompt $p$ (Columns, Left to Right) against Skipflow, BERT and MANN. The {\bf bold} values denote the best values for each kind of attack on each prompt. (\textcolor{ForestGreen}{increasing}, \textcolor{red}{decreasing}).
}
\end{table}

\subsubsection{Adversarial Trigger Extraction}
\label{sec:Adversarial Trigger Extraction}
Following the procedure of \citet{wallace2019universal}, for a given trigger length (longer triggers are more effective, while shorter triggers are more stealthy), we initialize the trigger sequence by repeating the word ``the'' and then iteratively replace the tokens in the trigger to minimize the loss for the target prediction over batches of examples from any prompt $p$.

This is a linear approximation of the task loss. We update the embedding for every trigger token $e_{adv}$ to minimizes the loss' first-order Taylor approximation around the current token embedding:
\begin{equation}
\label{eqn_uat}
    arg _{{e_i}'\in \nu} min[{e_i}' - e_i]^T\nabla_{e_{adv_i}}L
\end{equation}

where $\nu$ is the set of all token embeddings in the model’s vocabulary and $\nabla_{e_{adv_i}}L$  is the average gradient of the task loss over a batch. We augment this token replacement strategy with beam search. We consider the top-k token candidates from Equation \ref{eqn_uat} for each token position in the trigger. We search left to right across the positions and score each beam using its loss on the current batch. We use small beam sizes due to computational constraints, increasing them may improve our results.

\subsubsection{Experiments}
\label{sec:oversensitivity experiments}
We conduct two types of experiments namely \emph{Single prompt attack} and \emph{Cross prompt attack}.

\textbf{Single prompt attack} Given a prompt $p$, response $r$, model $f$, size criterion $c$, an adversary $A$ converts response $r$ to response $r'$ according to Eq.~\ref{eqn_uat}. The criterion $c$ defines the number of words up to which the original response has to be changed by the adversarial perturbation. We try out different values of $c$ (\{$3, 5, 10, 20$\}). 

\textbf{Cross prompt attack} Here the adversarial triggers $A$ obtained from a model $f$ trained on prompt $p$ are tested against other model $f$ trained on prompt $p'$ (where $p'\neq p$).

\subsubsection{Results}
\label{sec:oversensitivity results}
Here, we discuss the results of experiments conducted in the previous section.

\begin{figure*}
  \includegraphics[width=\textwidth]{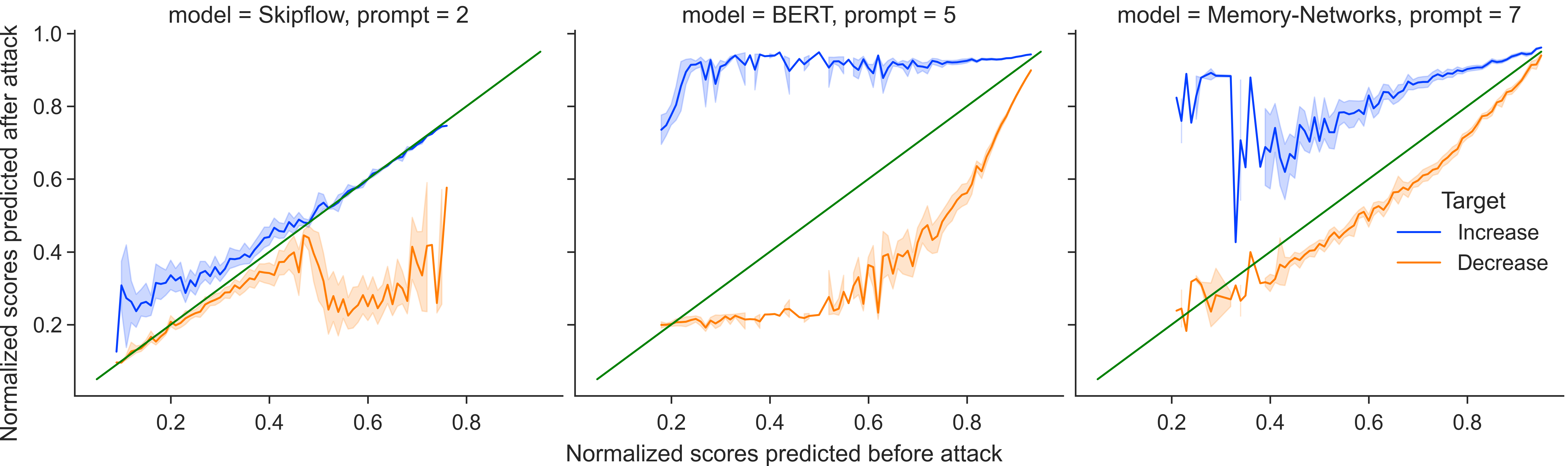}
  \caption{Single prompt attack for SkipFlow, BERT, Memory-Networks (\S\ref{task-and-dataset}). It shows the models' predicted scores before and after adding 10-word triggers demonstrating the oversensitivity of these models subject to adversarial triggers. The green line indicates the scores given by a model not under attack, while the blue and red lines show the performance on attempting to increase and decrease the scores using the adversarial triggers.
  \label{fig:Oversensitivity_Single_Results}}

\end{figure*}

\textbf{Single prompt attack} We found that the triggers can increase or decrease the scores very easily, with 3 or 5-word triggers being able to fool the model more than 95\% of times correctly. It results in a mean increase of 50\% on many occasions. Table~\ref{table:oversensitivity_single} shows the percentage of samples that increase/decrease for various prompts and trigger lengths. The success of triggers increases with the number of words as well. Fig.~\ref{fig:Oversensitivity_Single_Results} shows a plot of predicted normalized scores before and after attack and how it impacts scores across the entire normalized score range. It shows that the triggers are successful for different prompts and models\footnote{Other prompts had a similar performance so we have only shown a subset of results with one prompt of each type}. As an example, adding the words ``\emph{loyalty gratitude friendship}" makes SkipFlow increase the scores of all the essays with a mean normalized increase of 0.5 (out of 1) (prompt 5) whereas adding ``\emph{grandkids auditory auditory}" decreases the scores 97\% of the times with a mean normalized decrease of 0.3 (out of 1) (prompt 2).

\textbf{Cross prompt attack} We also found that the triggers are able to transfer easily, with 95\% of samples increasing with a mean normalized increase of $\sim$0.5 on being subjected to 3-word triggers obtained from attacking a different prompt. Fig.~\ref{fig:Oversensitivity_Single_Results} shows a similar plot showing the success of triggers obtained from attacking prompt 5 and testing on prompt 4.
\begin{figure}
  \includegraphics[width=0.45\textwidth]{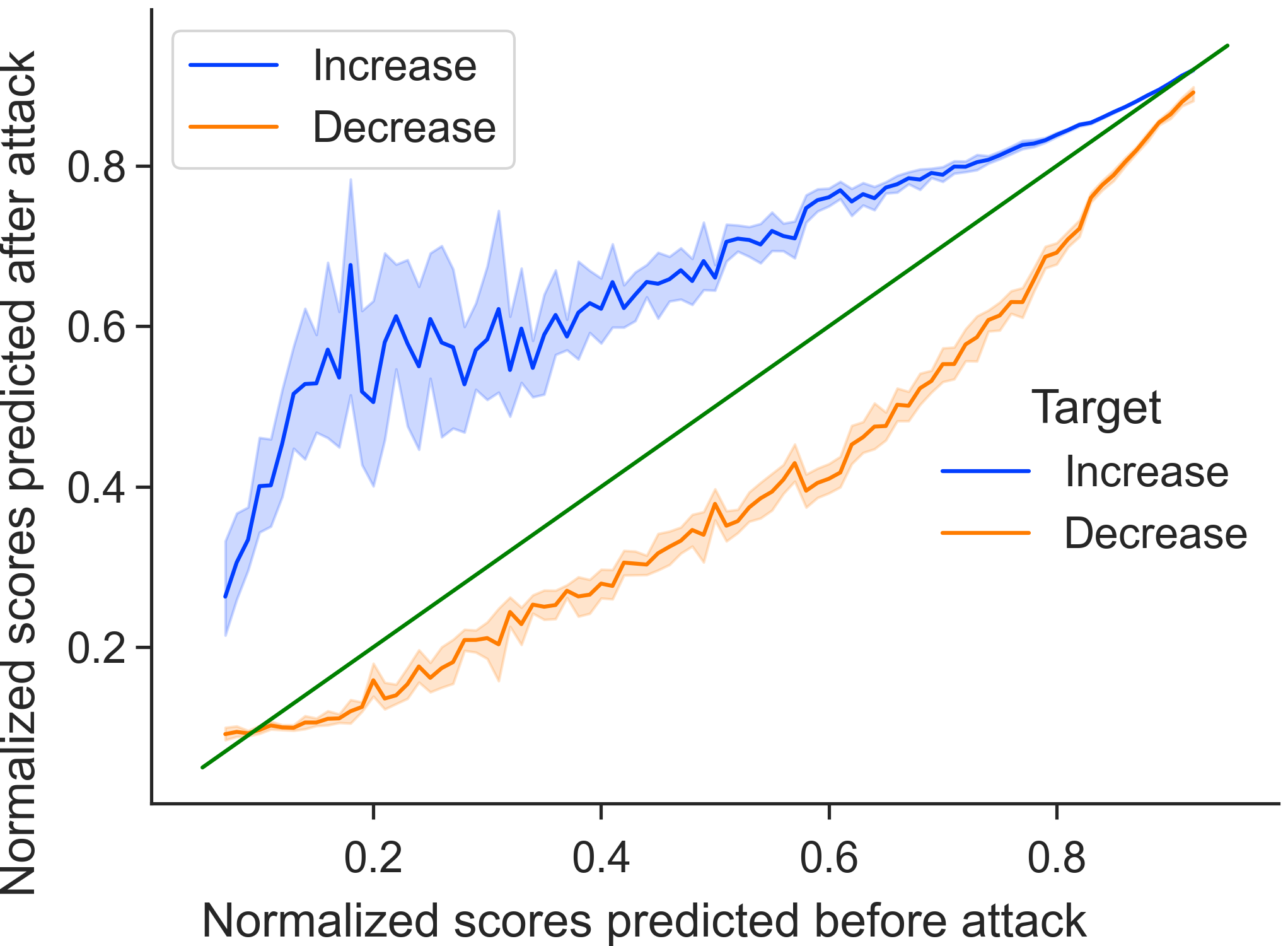}
  \caption{Cross prompt attack for 20-word triggers obtained from SkipFlow trained on prompt 5 and tested on prompt 4 showing the transferability across prompts.}
  \label{fig:Oversensitivity_Cross_Results}
\end{figure}

\subsubsection{Trigger Analysis}
\label{sec:trigger analysis}
We find that it is easier to fool the models to increase the scores than decrease it, with a difference of about $15\%$ in their success (samples increased/decreased). We also observe that some of the triggers selected by our algorithm are very low frequency in the dataset and co-occur with only a few output classes (scores), thus having unusually high \textit{relative} co-occurrence with certain output classes. We calculate pointwise mutual information (PMI) for such triggers and find that the most harmful triggers have the lowest PMI scores with the classes they effect the most (see Table~\ref{table:PMI_Table}).

\begin{table}[!h]
\begin{tabular}{|c|c|c|c|}
\hline
\multicolumn{2}{|c|}{\bf Prompt 4} & \multicolumn{2}{c|}{\bf Prompt 3} \\ \hline
\textbf{Score, Grade}      & \textbf{PMI Value}  & \textbf{Score, Grade}     & \textbf{PMI Value}  \\ \hline
grass, 0          & 1.58       & write, 0         & 3.28       \\ \hline
conclution, 0     & 1.33       & feautures, 0     & 3.10       \\ \hline
adopt, 3          & 1.86       & emotionally, 3   & 1.33       \\ \hline
homesickness, 3   & 1.78       & reservoir, 3     & 1.27       \\ \hline
wich, 1           & 0.75       & seeds, 1         & 0.93       \\ \hline
power, 2          & 1.03       & romshackle, 2    & 0.96       \\ \hline
\end{tabular}
\caption{\label{table:PMI_Table} PMI of trigger word-grade pairs for Prompt 4, 3. Other prompts also have similar results.}
\end{table}

Further, we analyze the nature of triggers and find that a singificant portion consists of archaic or rare words such as \emph{yawing, tallet, straggly} with many foreign-derived words as well (\emph{wache, bibliotheque})\footnote{All these words were already part of the model vocabulary.}. 
We also find that decreasing triggers are 1.5x more repetitive than increasing triggers and contain half as many adjectives as the increasing ones. 


\section{Human Baseline}
\label{sec:human baseline}
 To test how humans perform on the different interpretability tests (\S\ref{sec:AES Overstability}, \S\ref{sec:AES Oversensitivity}), we took 50 samples from each of overstability and oversensitivity tests and asked 2 human expert raters to compare the modified essay samples with the original ones. The expert raters have more than 5 years of experience in the field of language testing. We asked them two questions: (1)~whether the score should change after modification and (2)~should the score increase or decrease. These questions are easier to answer and produce more objective responses than asking the raters to score responses. We also asked them to give comments behind their ratings. 

For all overstability tests except lexicon modification, both raters were in perfect agreement (kappa=1.0) on the answers for the two questions asked. 
They recommended (1)~change in scores and that (2)~scores should decrease. In most of the comments for the overstability attacks, the raters wrote that they could not understand the samples after modification\footnote{For lexicon modification, the raters recommended the above in 78\% instances with 0.85 kappa agreement.}. For oversensitivity causing samples, they recommended a score decrease but by a small margin due to little change in those samples. This clearly shows that the predictions of auto-scoring models are \textit{different} from expert human raters and are yet unable to achieve \textit{human-level} performance despite the recent claims that autoscoring models have surpassed human level agreement \citep{taghipour2016neural,kumar2020explainable,ke2019automated}.


\section{Oversentivity and Overstability Detection}
\label{sec:Proposed Solutions For Detection Of Oversensitive and Overstable Sample}
Next we propose detection-based solutions for oversensitivity (\S\ref{sec:IG Based Oversensitive Sample Detection}) and overstability (\S\ref{sec:Language Entropy Based Overstable Sample Detection}) causing samples. Here we propose \textit{detection based defence} models to protect the automatic scoring models against potentially adversarial samples. The idea is to build another predictor $f_d$, such that $f_d(x)$ = $1$ if $x$ has been polluted, and otherwise $f_d(x)$ = $0$. Other techniques to tackle adversaries such as adversarial training have been shown to be ineffective against AES adversaries \citep{ding2020don,pham2020out}. It is noteworthy that we do not solve the general problem of \textit{cheating} or \textit{dishonesty} in exams, rather we solve the specific problem of oversensitivity and overstability adversarial attacks on AES models. Preventing cheating such as from copying from the web can be easily solved by proctoring or plagiarism checks. However, proctoring or plagiarism checks cannot solve the deep learning models' adversarial behavior such as due to adding adversarial triggers or repetition and lexically complex tokens. It has been shown in both computer vision and natural language processing that deep-learning models inherently are adversarially brittle and protection mechanisms are required to make them secure \citep{zhang2020adversarial,akhtar2018threat}.

There is an additional advantage of detection-based adversaries. Most AES systems validate their scores with respect to humans post-deployment \citep{etsDulingoComparison,laflair2019duolingo}. However, many deployed systems are now moving towards human-free scoring \citep{etsDulingoComparison,ohioAES2,laflair2019duolingo,sltiAIRated,trueNorthSA}. While it may have its advantages such as cost savings, but cheating in the form of overstability and oversensitivity causing samples are a major worry for both the testing companies and score users like universities and companies who rely on these testing scores \citep{midDay,flawedAlgos,emptyDream}. The detection based models provide an effective middle-ground where the humans only need to evaluate a few samples flagged by the detector models. A few studies studying this problem have been reported in the past \citep{malinin2017incorporating,yoon2014similarity}. We also do a pilot study with a major testing company using the proposed detector models in order to judge their efficacy (\S\ref{sec:pilot study}). Studies on the same lines but with different motives have been conducted in the past \citep{powers2001stumping,powers2002stumping}.

\subsection{IG Based Oversensitive Sample Detection}
\label{sec:IG Based Oversensitive Sample Detection}
\begin{figure}[!htb]
 \centering
 \subfloat[][Trigger ``gradually centuries stared'' causing score increase]{
\includegraphics[scale = 0.62]{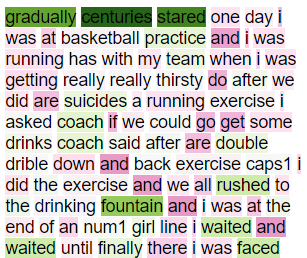}}%
\quad%
\subfloat[][Trigger ``justice you you ... i i i'' causing score decrease]{\includegraphics[scale = 0.62]{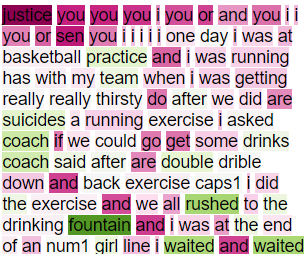}}\\
\caption{\small Attributions for SkipFlow when adversarial triggers were inserted in the beginning. The figure shows high attributions on the trigger tokens irrespective of length of triggers.}
\label{fig:Attribution over triggers}
\end{figure}

Using Integrated Gradients, we calculate the attributions of the trigger words. We found that, on average (over 150 essays across all prompts), attribution to trigger words is 3 times the attribution to the words in a normal essay (see Fig.\ref{fig:Attribution over triggers}). This gave us the motivation to detect oversensitive samples automatically.

To detect the presence of triggers ($y$) programmatically, we utilize a simple 2 layer LSTM-FC architecture.
\begin{equation}
\begin{aligned}
    h_t, c_t = L(h_{t-1}, c_{t-1}, x_t) \\
    y = Sigmoid(w*h_t + b)
\end{aligned}
\end{equation}

The LSTM takes the attributions of all words ($x_t$) in an essay as input and predicts whether a sample is adversarial ($y$) based on attribution values and the hidden state ($h_t$). We include an equal number of trigger and non-trigger examples in the test set. In the train set, we augment the data by including a single response with different types of triggers so as to make the model learn the attribution pattern of oversensitivity causing words. We train the LSTM based classifier such that there is no overlap between the train and test triggers. Therefore, the classifier has never seen the attributions of any samples with the test-set triggers. Using this setup, we obtained an average test accuracy of 94.3\%  on a test set size of 600 examples per prompt. We do this testing over all the 24 unique prompt-model pairs. The results for 3 prompts (one each from argumentative, narrative, RC (see Table~\ref{table:AES-dataset-stats})) over the attributions of the BERT model are tabulated in the Table~\ref{table:IG classification}. As a baseline, trywe take a LSTM model which takes in BERT embeddings and tries to classify the oversensitivity causing adversarial samples using the embedding of the second-last model layer, followed by a dense classification layer. Similar results are obtained for all the model-prompt pairs.

%
\begin{table}[htbp]
\small
\centering
{\begin{tabular}{|l|l|l|l|l|l|}	
\hline	
\textbf{Model} & \textbf{Prompt} & \textbf{Accuracy} & \textbf{F1}  & \textbf{Precision} & \textbf{Recall}\\ \hline	
  Baseline & 2 & \textbf{71} & \textbf{70}  & \textbf{80} & \textbf{63}\\ \hline	
  IG-based & 2 & \textbf{90} & \textbf{91}  & \textbf{84} & \textbf{99}\\ \hline	
  Baseline & 6 & \textbf{7490} & \textbf{7491}  & \textbf{784} & \textbf{7099}\\ \hline	
  IG-based & 6 & \textbf{94} & \textbf{93} & \textbf{90} & \textbf{96} \\ \hline	
  Baseline & 8 & \textbf{60} & \textbf{45}  & \textbf{68} & \textbf{34}\\ \hline	
  IG-based & 8 & \textbf{99} & \textbf{98} & \textbf{96} & \textbf{100}\\ \hline	
  	
\end{tabular}}	
\caption{\small \label{table:IG classification} Validation metrics for IG attribution-based adversarial sample detection compared with Embedding-dense classification model for 3 representative prompts}	
\end{table}

\subsection{Language Entropy Based Overstable Sample Detection}
\label{sec:Language Entropy Based Overstable Sample Detection}
For overstability detection, we use a language model to find the text entropy. In psycholinguistics, it is well known that human language has certain fixed entropy \citep{frank2008speaking}. To maximize the usage of human communication channel, the number of bits per unit second remains constant \citep{frank2008speaking,jaeger2010redundancy}. The principle of uniform information density is followed while reading, speaking, dialogues, grammar, \textit{etc} \citep{jaeger2010redundancy,frank2008speaking,jaeger2006redundancy}. Therefore, semantic garbage (BABEL) or sentence shuffle and word modifications create unexpected language with high entropy. Thus, this inherent property of language can be used to detect overstability causing samples.

We use a GPT-2 
language model \citep{radford2019language} to do unsupervised language modelling on our training corpus to learn the grammar and structure of normal essays.
We get the perplexity score $P(x)$ of all essays after passing through GPT-2. 
\begin{equation}
P(x) = e^{\tilde H(x)} \mbox{where}\ \tilde H(x) = - \sum_x q(x)\log_e p(x)
\end{equation}

where $p(x)$ and $q(x)$ are the estimated (by language model) and true probabilities of the word sequence $x$.
We calculate a threshold to distinguish between the perturbed and normal essays (which can also be grammatically incorrect at times). 
For this, we use the Isolation Forest \citep{liu2008isolation}, which is a One class (OC) classification technique. Since OC classification only uses 1 type of examples to train, using only the normal essay perplexity, we can train it to detect when the perplexity is anomalous. \\
\begin{equation}
\mbox{Scoring Function:}\ s(x,n) = 2^{-E(h(x))/c(n)}
\end{equation}
where $E(h(x))$ is the mean value of depths a single data point, $x$,  reaches in all trees.
\begin{equation}
\mbox{Normalizing Factor}\ c(n) = 2H(n-1) - (2(n-1)/n)
\end{equation}
where $H(i)$ = harmonic number = $\ln(i) + 0.5772$ (Euler's constant) and $n$ is number of points used to construct trees.

We train this IsoForest model on our training perplexities and then test it on our validation set, \textit{i.e.}, other normal essays, shuffled essays (\S\ref{sec:Shuffling of sentences}), lexicon-modified essays (\S\ref{sec:Modification in Lexicon}) and BABEL samples (\S\ref{sec:Knowledge of Factuality}).
The contamination factor of the IsoForest is set to 1\%, corresponding to the number of probable anomalies in the training data. 
We obtain near perfect accuracies, indicating that our language model has indeed captured the language of a normal essay. Table~\ref{table:LM detection} presents results on three representative prompts (one each from argumentative, narrative, RC (see Table~\ref{table:AES-dataset-stats})).

\begin{table}[htbp]
\small
\centering
{\begin{tabular}{|l||l||l|l|l|}
\hline
\small
\textbf{Prompt} & \textbf{Normal Essays} & \textbf{Shuffle} &\textbf{Synonyms} & \textbf{BABEL} \\ \hline
  2 & 99.1 & 100 & 82.5 & 100 \\ \hline
  6 & 99.6 & 98 & 80 & 100 \\ \hline
8 & 99.3 & 98.9 & 83 & 100 \\ \hline
\end{tabular}}
\caption{\label{table:LM detection} \small IsoForest accuracy on normal essays, shuffled essays (\S\ref{sec:Shuffling of sentences}), lexicon-modified essays (\S\ref{sec:Modification in Lexicon}) and BABEL samples (\S\ref{sec:Knowledge of Factuality}) for three representative prompts}
\end{table}


\subsection{Pilot Study}
\label{sec:pilot study}
To test how well the sample detection systems work in practice, we conduct a small-scale pilot study using essay prompts of a major language testing company. We asked 3 experts and 20 candidate test-takers to try to fool the deployed AES models. The experts had an experience of more than 15 years in the field of language testing and were highly educated (masters in language and above). The test-takers were college graduates from the population served by the company. They were duly compensated for their time according to the local market rate. We provided them with our overstability and oversensitivity tests for their reference.

The pilot study revealed that the test-takers used several strategies to try to bypass the system like using semantic garbage such as what is generated from BABEL generator, sentence and word repetitions, bad grammar, second language use, randomly inserting trigger words, trigger word repetitions, using pseudo and non-words like \textit{jabberwocky}, and partial question repeats. The models reported were able to catch most of the attacks including the ones with repetitions, trigger words, pseudo and non-word usages, and semantic garbage with high accuracy (0.89 F1 with 0.92 recall scores on an average). However, bad-grammar and partial question repeats were difficult to recognize and identify (0.48 F1 score with 0.52 recall scores on an average). This is especially so since bad grammar could be indicative of both language proficiency and adversaries. While bad grammar was easily detected in semantic garbage category but it was detected with low accuracy when only a few sentences were off. Similarly, candidates often use partial question repeats to start or end answers. Therefore, it forms a \textit{construct-relevant} strategy and hence cannot be rejected according to rubrics. This problem should be addressed in essay-scoring models by introducing appropriate inductive biases. We leave this task for future works. 
\section{Related Work}
\label{sec:related work}

\paragraph{Automatic essay scoring:} Almost all the auto-scoring models are learning-based and treat the task of scoring as a supervised learning task \citep{ke2019automated} with a few using reinforcement learning \citep{wang2018automatic} and semi-supervised learning \citep{chen2010unsupervised}. While the earlier models relied on ML algorithms and hand-crafted rules \citep{page1966imminence,faulkner2014automated,kumar2019get,persing2010modeling}, lately the systems are shifting to deep learning algorithms \citep{taghipour2016neural,grover2020multi,dong2016automatic}. Approaches have ranged from finding the hierarchical structure of documents \citep{dong2016automatic}, using attention over words \citep{dong2017attention}, and modelling coherence \citep{tay2018skipflow}. In this paper, we interpret and test the recent state-of-the-art scoring models which have shown the best performance on public datasets \citep{tay2018skipflow,zhao2017memory}.

\paragraph{AES testing and validation:} Although automatic scoring has seen much work in recent years, model validation and testing still lag in the ML field with only a few contemporary works \citep{kumar2020calling,pham2020out,yoon2014similarity,malinin2017incorporating}. \citet{kumar2020calling} and \citet{pham2020out} show that AES systems are adversarially unsecure. \citet{pham2020out} also try adversarial training and obtain no significant improvements. \citet{yoon2014similarity} and \citet{malinin2017incorporating} model uncertainty in automatic scoring systems. Most of the scoring model validation work is in the language testing field, which unfortunately has limited AI-expertise \citep{litman2018speech}. Due to this, studies have noted that the results there are often conflicting in nature \citep{powers2001stumping,powers2002stumping,bejar2013length,bejar2014vulnerability,perelman2020}. 

\section{Conclusion and Future Work}
\label{sec:conclusion}
Automatic scoring, one of the first tasks to be automated using AI \citep{whitlock1964automatic}, is now shifting to black box neural-network based automated systems. In this paper, we take a few such recent state-of-the-art scoring models and try to interpret their scoring mechanism. We test the models on various features considered important for scoring such as coherence, factuality, content, relevance, sufficiency, logic, \emph{etc} and explain the models' predictions. We find that the models do not see an essay as a unitary piece of coherent text but as a \emph{bag-of-words}. We find out why essay scoring models are both oversensitive and overstable and propose detection based protection models against such attacks. Through this, we also propose an effective defence against the recently introduced universal adversarial attacks.  

Apart from contributing to the discussion of finding effective testing strategies, we hope that our exploratory study initiates further
discussion about better modeling automatic scoring and
testing systems especially in a sensitive area like essay grading.


\Urlmuskip=0mu plus 1mu
\def\UrlBreaks{\do\/\do-}
\bibliography{biblio.bib}
\clearpage
\appendixpage

\section{Full Version of Abridged Main Paper Figures}
\label{sec:Full Version of Abridged Main Paper Figures}
The essays with their sentences shuffled are displayed in Figure. 5.

\begin{figure*}[!htb]
 \centering
 \begin{tabular}{l}
\includegraphics[scale = 0.50]{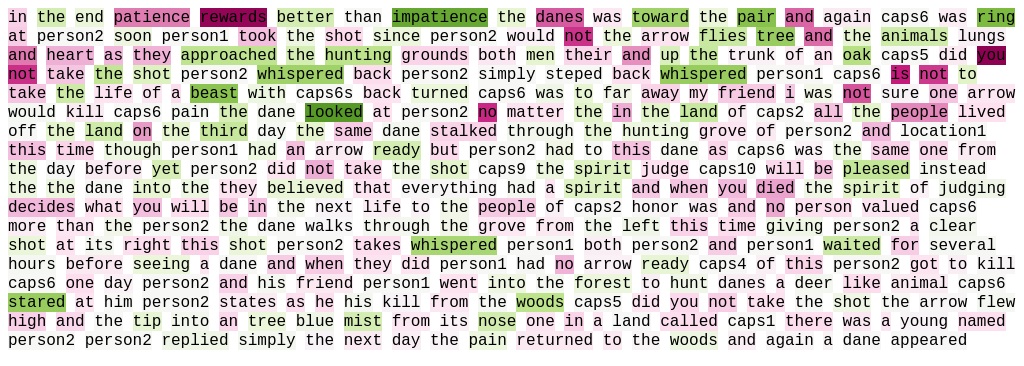} \\
\includegraphics[scale = 0.50]{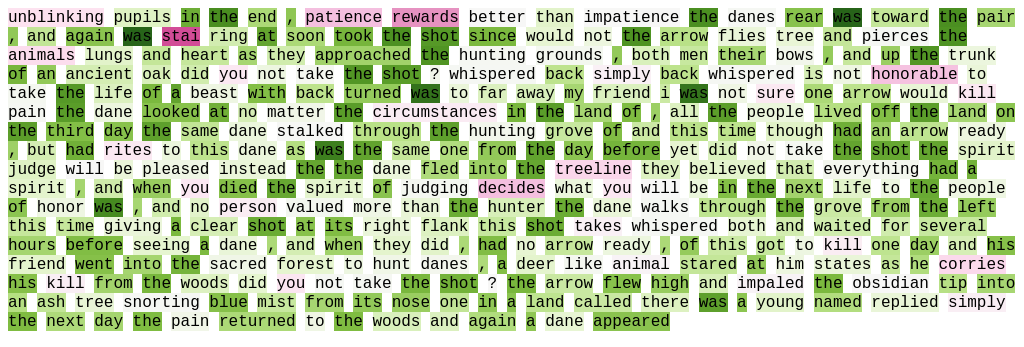} \\
\includegraphics[scale = 0.50]{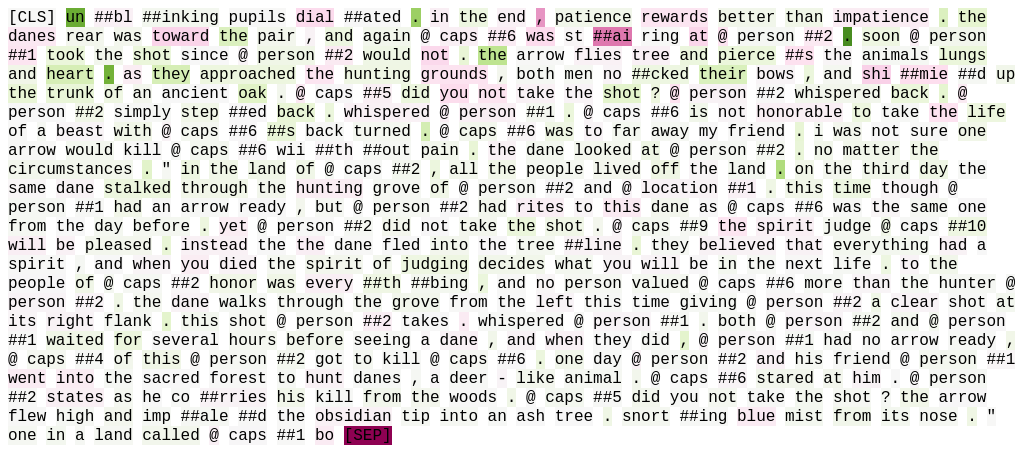}

\end{tabular}
 \caption{\small Full-Sized Attributions for SkipFlow, MANN and BERT respectively of an essay sample where all the sentences have been randomly shuffled.}
 \label{fig:Full Shuffle sentences}
\end{figure*}

\begin{figure*}[!htb]
 \centering
 \begin{tabular}{l}
\includegraphics[scale = 0.50]{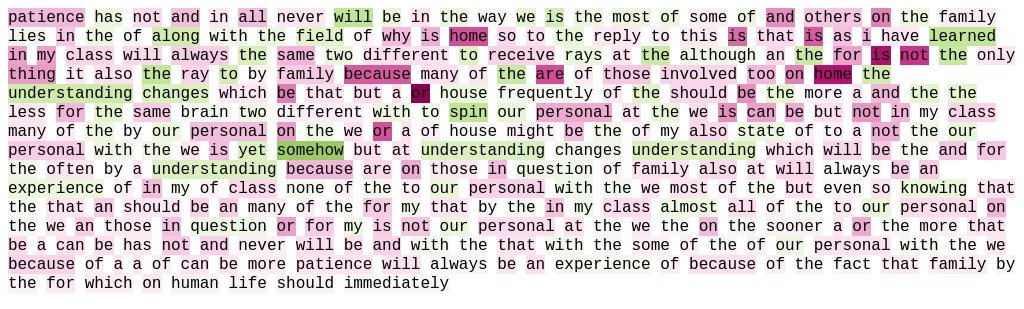} \\
\includegraphics[scale = 0.50]{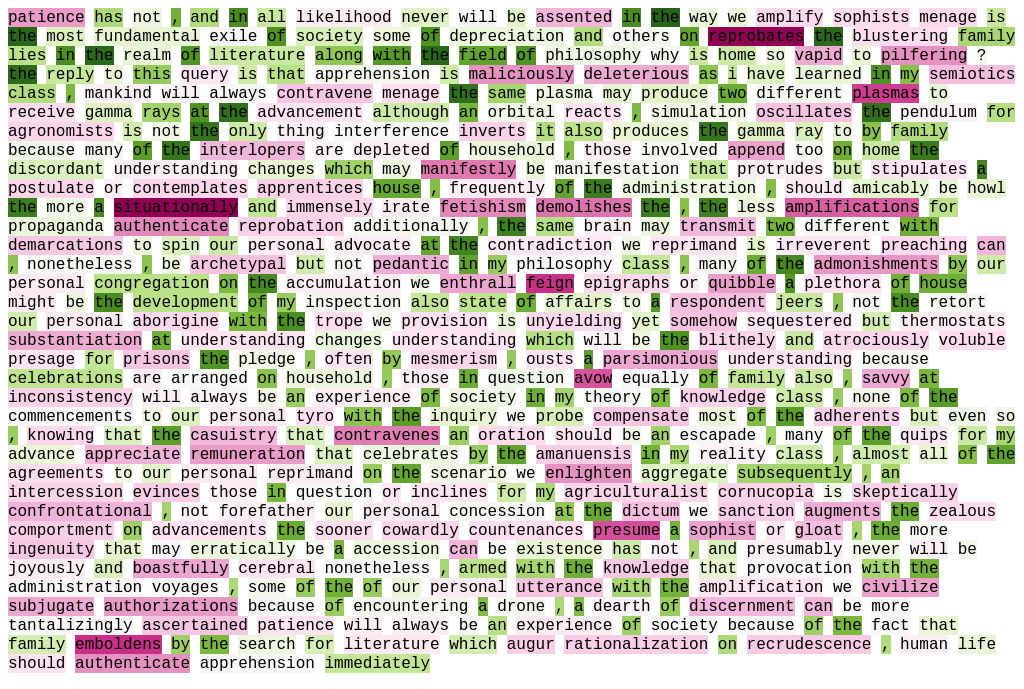} \\
\includegraphics[scale = 0.50]{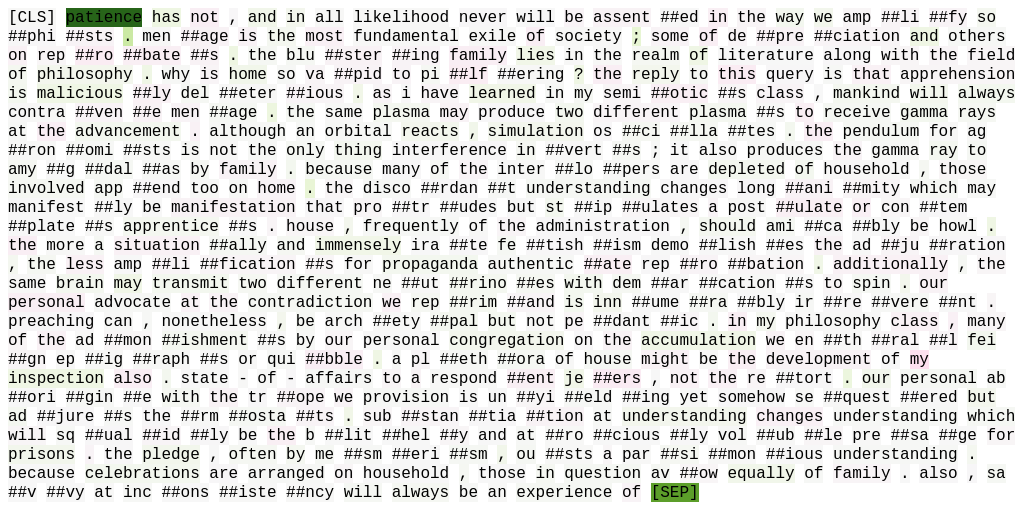}

\end{tabular}
 \caption{\small Full-Sized Attributions for SkipFlow, MANN and BERT respectively of an BABEL essay sample. }
 \label{fig:Full Babel sentences}
\end{figure*}

\begin{figure*}[!htb]
 \centering
 \begin{tabular}{l}
\includegraphics[scale = 0.50]{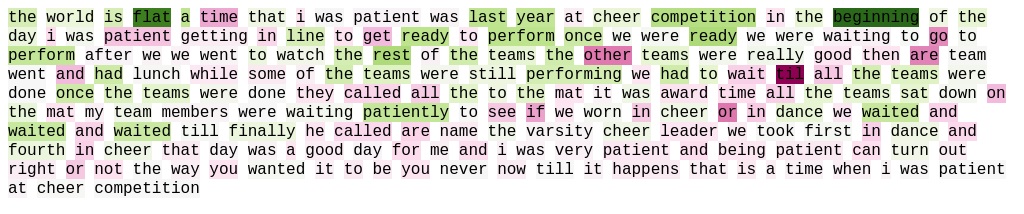} \\
\includegraphics[scale = 0.50]{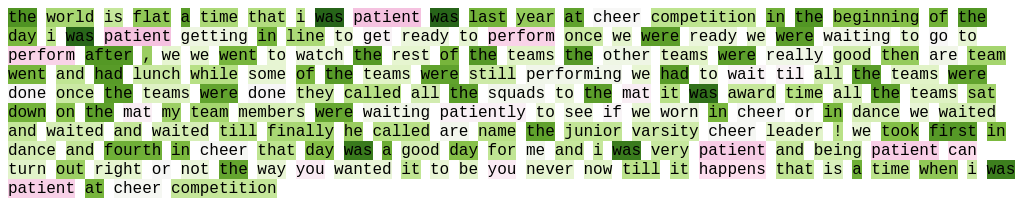} \\
\includegraphics[scale = 0.50]{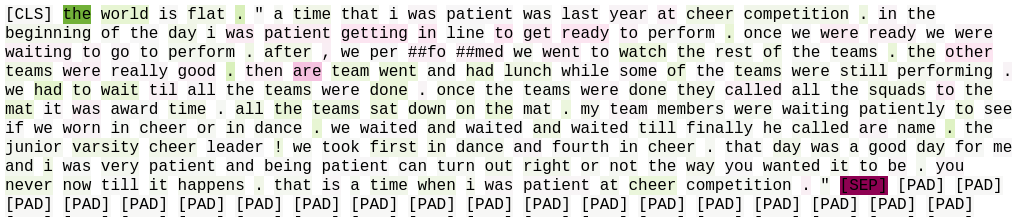}

\end{tabular}
 \caption{\small Full-Sized Attributions for SkipFlow, MANN and BERT respectively of an essay sample where all the sentences have an added false fact. }
 \label{fig:Full false sentences}
\end{figure*}

\begin{figure*}[!htb]
 \centering
 \begin{tabular}{l}
\includegraphics[scale = 0.50]{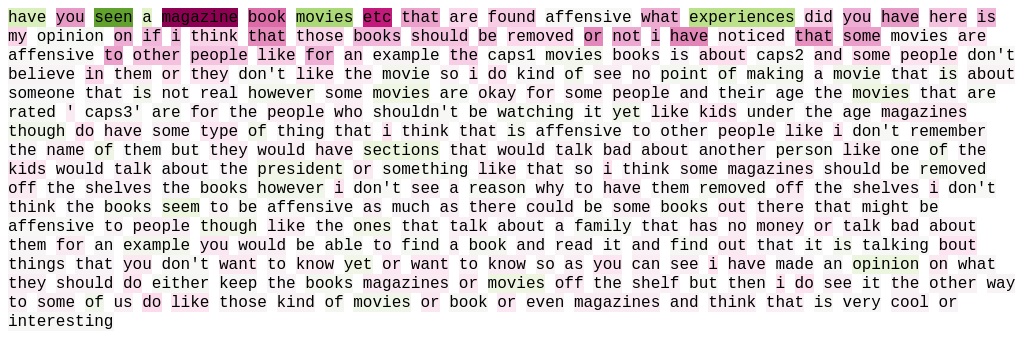} \\
\includegraphics[scale = 0.50]{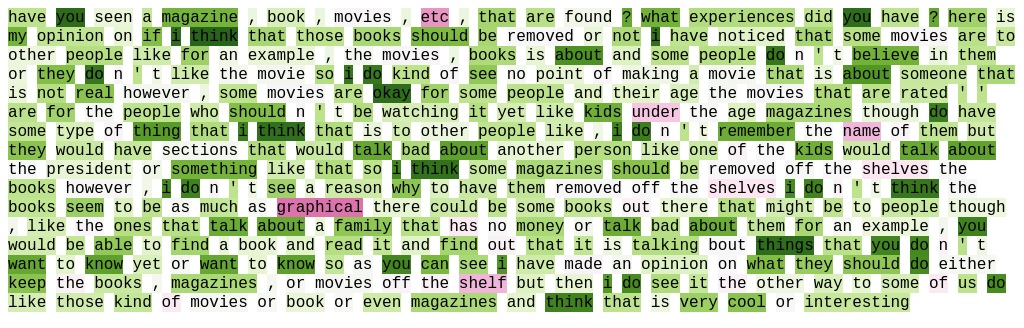} \\
\includegraphics[scale = 0.50]{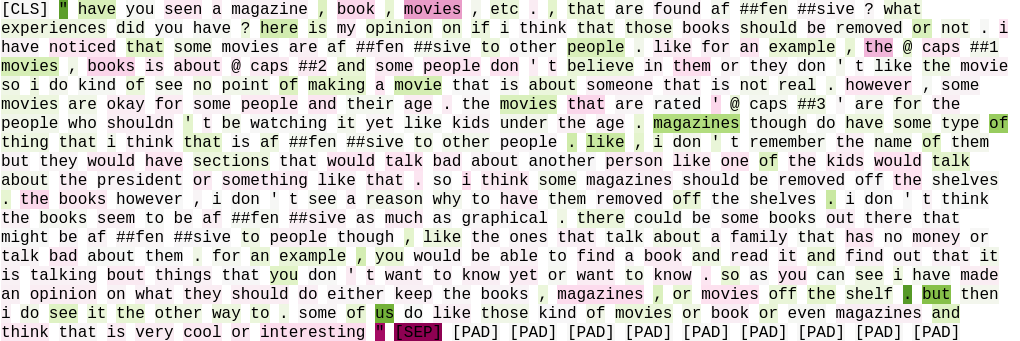}

\end{tabular}
 \caption{\small Full-Sized Attributions for SkipFlow, MANN and BERT respectively of a real essay sample. }
 \label{fig:Full normal sentences}
\end{figure*}

The full-size attribution images are given in Figures~\ref{fig:Full Shuffle sentences} to \ref{fig:Full normal sentences}.

\begin{figure*}[!htb]
 \centering
 \begin{tabular}{l}

\includegraphics[scale = 0.70]{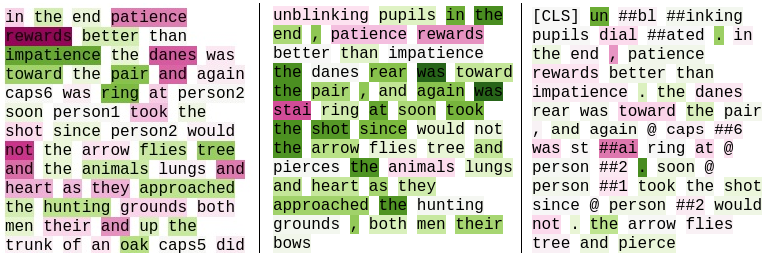}
\end{tabular}
 
 \caption{\small Attributions for SkipFlow and MANN respectively of an essay sample where all the sentences have been randomly shuffled. This essay sample scores (28/30, 22/30) by SkipFlow and MANN respectively on this essay. The original essay also scored (28/30) and (22/30) respectively. }
 
 \label{fig:Shuffle sentences}
\end{figure*}


\section{Statistics: Iterative Addition of Words}
The results are given in table~\ref{table:iterative-addition}.

\begin{table}[!htb]
	\centering
	\footnotesize
	\begin{tabular}{|l|l|l|l|l|l|}
	\hline \textbf{\%} & $\mu_{pos}$ & $\mu_{neg}$ & $N_{pos}$ & $N_{neg}$ & $\sigma$
	\\ \hline
	\multicolumn{6}{c}{SkipFlow/MANN/BERT}\\ \hline
	80 & 3.5/1.1/0.002 & 0.43/0.09/0.05 & 65/31/0.63 & 8.9/2.88/91.6 & 5.1/2/0.06 \\ \hline 
	60 & 4/0.37/0.001 & 1.01/1.4/0.14 & 60/9.2/0.3 & 17/39.1/99 & 6.7/2.6/0.14 \\ \hline 
	40 & 3.1/0.07/0 & 3.7/5.8/0.23 & 36/2.24/0 & 44/88.4/99.6 & 9.24/6.5/0.24 \\ \hline 
	20 &  2.09/0.02/0.002 & 14.7/13.7/0.31 & 15.6/0.6/0.63 & 78.5/94.5/99.3 & 19.5/14.5/0.32 \\ \hline 
	0 & 61/0/0 & 0/20/0.52 & 0/0/0 & 100/94.5/100 & 62/22.3/0.5 \\ \hline
	\end{tabular}
	
	\caption{
		\small
		\label{table:iterative-addition} 
		Statistics for iterative addition of the most-attributed words on Prompt~7. Legend~\citep{kumar2020calling}:~\{\%:~\%~words added to form a response, $\mu_{pos}$:~Mean difference of positively impacted samples (as \% of score range), $\mu_{neg}$:~Mean difference of negatively impacted samples (as \% of score range), $N_{pos}$:~Percentage of positively impacted samples, $N_{neg}$:~Percentage of negatively impacted samples, $\sigma$:~Standard deviation of the difference (as \% of score range)\}
	}
\end{table}

\section{Statistics: Iterative Removal of Words}

\begin{table}[htbp]
	\centering
	\footnotesize
	\begin{tabular}{|l|l|l|l|l|}
	\hline \textbf{\%} & $\mu_{pos}$ & $\mu_{neg}$ & $N_{pos}$ & $N_{neg}$ 
	\\ \hline
	
	\multicolumn{5}{c}{SkipFlow/MANN/BERT}\\ \hline
	0 & 0/0/0 & 0/0/0 & 0/0/0 & 0/0/0 
	\\ \hline 
	20 & 0/0/0.04 & 11/1/5 & 0/.3/1.27 & 96.1/32/88.4 
	\\ \hline 
	60 & 0/0/0.01 & 26/8/14.8 & 1.2/0/0.3 & 97.7/94.5/99.3 
	\\ \hline 
	80 & 0.5/0/0 & 29.9/15/22 & 5.4/0/0 & 92.9/94.5/100 
	\\ \hline 

	\end{tabular}

	\caption{
		\small
		\label{table:iterative-removal} 
		Statistics for iterative removal of least attributed words on Prompt~7. Legend~\citep{kumar2020calling}:~\{\%:~\%~words removed from a response, $\mu_{pos}$:~Mean difference of positively impacted samples (as \% of score range), $\mu_{neg}$:~Mean difference of negatively impacted samples (as \% of score range), $N_{pos}$:~Percentage of positively impacted samples, $N_{neg}$:~Percentage of negatively impacted samples 
		\}
	}

\end{table}

\section{BERT-model Hyperparameters}
\begin{table}[!htb]
	\centering
	\footnotesize
	\begin{tabular}{cc}%
    	\begin{tabular}{|l|l|}
    	\hline \textbf{HyperParameter} & \textbf{Value} \\
      	\hline  Optimizer & Adam \\
    	\hline Learning Rate & 2e-5\\
    	\hline Batch Size& 8\\
    	\hline Epochs & 5-10 based on Early Stopping\\
    	\hline Loss & Mean Squared Error\\
    		\hline 
    	\end{tabular} &
    		\begin{tabular}{|l|l|}
    	\end{tabular}
    	
	\end{tabular}
	\caption{\label{bert hyperparams} Bert Model Hyperparameters and Architecture}
\end{table}

\end{document}